# AN AUTOMATED MULTIPLE-CHOICE QUESTION GENERATION USING NATURAL LANGUAGE PROCESSING TECHNIQUES

Chidinma A. Nwafor[1] and Ikechukwu E. Onyenwe[2]

[1]Department of Computer Science, Nnamdi Azikiwe University Awka, Anambra State, Nigeria
[1]nwaforchidinmaa@gmail.com and [2]ie.onyenwe@unizik.edu.ng

## ABSTRACT

*Automatic multiple-choice question generation (MCQG) is a useful yet challenging task in Natural Language Processing (NLP). It is the task of automatic generation of correct and relevant questions from textual data. Despite its usefulness, manually creating sizeable, meaningful and relevant questions is a time-consuming and challenging task for teachers. In this paper, we present an NLP-based system for automatic MCQG for Computer-Based Testing Examination (CBTE). We used NLP technique to extract keywords that are important words in a given lesson material. To validate that the system is not perverse, five lesson materials were used to check the effectiveness and efficiency of the system. The manually extracted keywords by the teacher were compared to the auto-generated keywords and the result shows that the system was capable of extracting keywords from lesson materials in setting examinable questions. This outcome is presented in a user-friendly interface for easy accessibility.*

## 1. INTRODUCTION

According to [1], automatic question generation (AQG) is the task of automatically generating syntactically sound, semantically correct and relevant questions from various input formats such as text, a structured database or a knowledge base. AQG can be naturally applied in many domains such as Massive Open Online Course (MOOC), setting objective/subjective questions, automated help systems, search engines, Chabot systems (e.g. for customer interaction), and healthcare for analyzing mental health [2]. Despite its usefulness, manually creating meaningful and relevant questions is a time-consuming and challenging task. For example, while evaluating students on reading comprehension, it is tedious for a teacher to manually create questions, find answers to those questions, and thereafter evaluates answers.

In the field of education, questioning is widely acknowledged as an effective instructional strategy to evaluation or assessment of students or the learner at the end of a lesson. Questioning as an instructional tool can be traced back to the fourth century, when Socrates used questions and answers to challenge assumptions, expose contradictions, and lead to new knowledge and wisdom. Since Socrates, questioning was employed to encourage students to use higher-order thinking processes and it is the dominant mode of teacher-student interaction accounted for almost 80% of the total interactions [3]. There are three major types of assessment for evaluating students' learning capabilities – Formative assessment, Interim assessment and Summative assessment. Generating questions that covers most areas of a lesson note would help in Formative and Interim assessment because the Formative assessment helps to improve students' understanding and performance while the Interim assessment helps teachers identify gaps in student understanding and instruction. Multiple-choice Question (MCQ)s that widely cover all the areas of a lesson note is an important effort to test students on the above assessments and the students' overall cognitive levels held at the end of a lesson. Today, MCQs in the area of Computer-Based Testing Examination (CBTE) is the most commonly used tool for assessing the knowledge capabilities of learners in the field of learning.

Properly constructed MCQs can assess higher cognitive processing of Bloom's taxonomy such as interpretation, synthesis and application of knowledge, instead of just testing recall of isolated facts. Most institutions and Centre of learning have embraced MCQ testing as the foundation of their testing systems, thereby saving cost especially where large numbers of candidates are involved—and it is a format that can provide precision where other measurement options may be lacking such as observing performance.

Effective style of questioning as described by [4] is always an issue to help students attend to the desired learning outcome. Furthermore, to make it effective, balancing between lower and higher level question is a must [4]. Time-consuming factor in setting MCQs hinders achieving the above assessments and good learning outcomes due to MCQs generated tend to cover limited areas of a lesson note from which the MCQs are taken from. Therefore, it is necessary for an automatic system that will bootstrap the existing method of setting MCQs.

Natural Language Processing (NLP) is a branch of artificial intelligence that deals with the interaction between computers and humans using the natural language. The ultimate objective of NLP is to read, decipher, understand, and make sense of the written or textual data in natural languages of human in a manner that is valuable. In this paper, we focus on the setting of MCQs for CBTE through NLP to improve the method of setting MCQs and modification, and for creating a viable question bank for subsequent use by the academicians for their learners. This will ensure an MCQ that includes appropriate questions and options based on the learning objectives and importance of the topics discussed in a lesson material. We used NLP methods, Term Frequency-Inverse Document Frequency (TF-IDF) and N-grams, to extract the most significant words present in a lesson material and the selection does not depend on any vocabulary; these words are keywords that capture the main topics discussed in a lesson material, they also serve as an indication of document relevance for users in an information retrieval (IR) environment.

## 2. RELATED LITERATURE

[5] developed ArikIturri, an AQG for Basque language test questions. The information source for this question generator consists of linguistically analysed real corpora, represented in XML mark-up language. [6] researched on using NLP on narrative texts to find discourse connectives for AQG. Discourse connectives are words or phrases that indicate relationships between two logical sentences or phrases and suggest the presence of mutually related extended verbal expression. The questions were generated by first extracting the text from the materials supplied by the user using text processing concept of NLP. [7] developed a system which can generate various logical questions from the given text input. The system uses 3-step strategy, viz; Select the best potential set of sentences from the given text input from which we could generate the questions, find the subject and context of the sentence to find its core agenda (Gap Selection) and analyze the best form of question that can be generated from that sentence (Question Formation). [8] proposed a method for automatically generating distractors for multiple-choice English vocabulary questions. The proposed method introduces new sources for collecting distractor candidates and utilises semantic similarity and collocation information when ranking the collected candidates. [9] looked at the development of an automatic factual open cloze question generation system which can generate fill-in-the-blank questions without alternatives. The system first extracts a set of informative sentences from the given input corpus based on Part-of-Speech tagging based rules.

Then, generate questions by omitting the answer-keys which are selected by identifying domain specific words in the sentences. The system also suggests answer hints for the examinees to reduce the number of possible answers that make assessment easier.

## 3. EXPERIMENTAL METHODOLOGY

The paper is based on the generation of MCQG through Natural Language Processing Techniques.

This is used to process the lesson materials/documents fed by the teacher into a multi-choice questions alongside the answers to each questions. The NLP processes are applied using TF-IDF and N-gram. This system only applies to the text of the document or lesson material in extracting keywords presented by the teacher. The document is converted into text file, loaded into the system memory and stored on the system. The text is split into sentences. The split sentences are tokenized, from which the corpus are built as TF-IDF and N-gram mode. From Figure 1, the raw data is the original text file from the document (lesson material). Noise removal is any piece of text which is not relevant to the context of the data such as stop words. Word Normalization converts high dimensional features or N different features to the low dimensional space or 1 feature, it includes tokenization and stemming. The cleaned text is the end result ready for further experimentation and such is used to analyse and auto-generate multi-choice questions (MCQs). Furthermore, we develop an application interface for easy access and presentation of the MCQs to the teachers for acceptance or rejection.

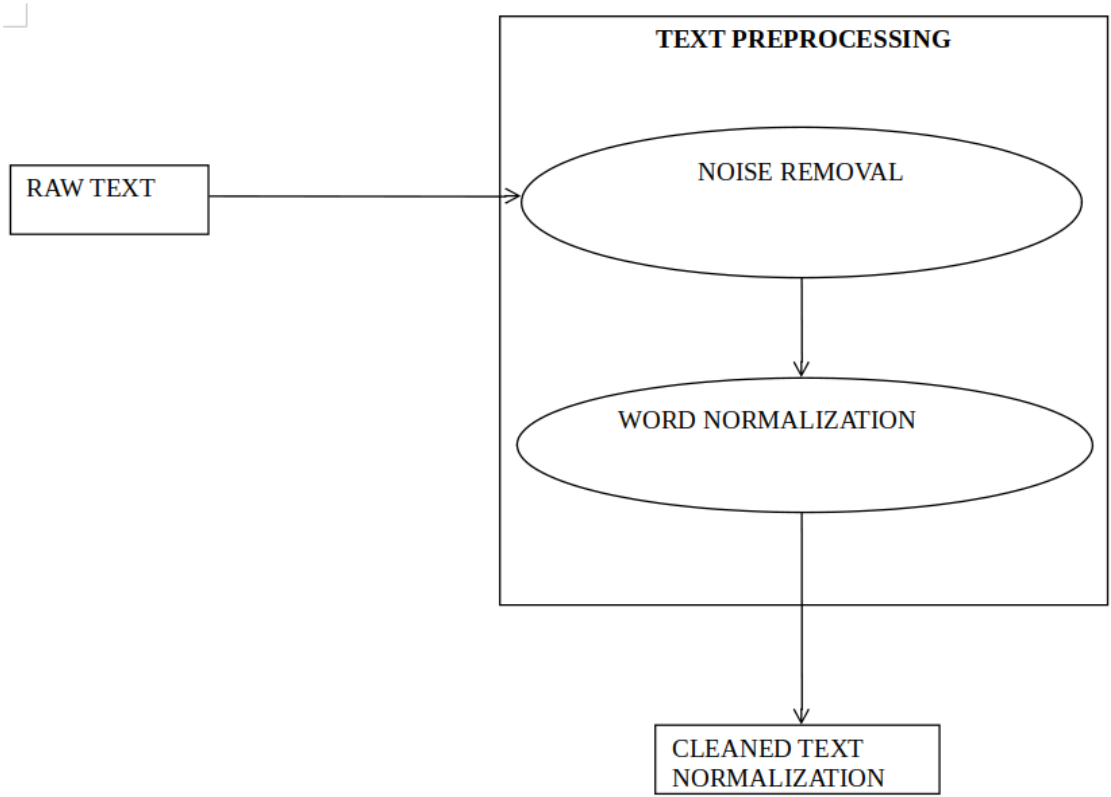

Figure 1. Text Preprocessing of Input Data.

## 3.1. Dataset and Preparation

In other to compute the performance evaluation of this paper to help check that it is not perverse, keywords extracted by the NLP techniques for accuracy computation was validated. First, five different lecture materials of varying sizes of sentences was selected (see Table 1).

Table1. Statistics of the Textual Data Used

| Lesson Materials | Number of Sentences | Number of Words | Number of Keywords by Teachers |
|---|---|---|---|
| Computer Science | 39 | 956 | 206 |
| Geology | 33 | 903 | 86 |
| Philosophy | 46 | 820 | 100 |
| Religion | 61 | 1474 | 187 |
| History | 64 | 1472 | 210 |

In order to evaluate the accuracy of the keywords extraction methods on these lesson materials, it was necessary to collect manually marked keywords to create a gold standard for the evaluation. The usual procedure of creating a gold standard of keywords is having one professional indexer who extracts the keywords based on his expertise and training, but also based on a set of guidelines defined by the interested parties. However, for this experiment, it was felt that the teachers (T) are appropriate for setting a gold standard. They know what words are most relevant in their lecture materials. Each lecture materials were annotated by a T to set a gold standard to which each set of auto-generated keywords could be compared. The statistics of the data used are stated in Table 1.

Table 2. Descriptive statistics of word counts.

| Lesson Materials | Minimum Number of Sentences | Maximum Number of Words | Mean number of Sentence |
|---|---|---|---|
| Computer Science | 6.0 | 59.0 | 24.5 |
| Geology | 10.0 | 43.0 | 25.8 |
| Philosophy | 1.0 | 39.0 | 17.8 |
| Religion | 1.0 | 59.0 | 24.2 |
| History | 1.0 | 79.0 | 23.0 |

Table 2 explains the columns "number of Sentences" and "number of words" in Table 1. For instance
Computer Science in Table 3.2, the least sentence has 6-word length of size while maximum sentence has 59-word length of size. Summation of all the sizes of the sentences divided by the total number of sentences gives the mean size of sentences in Computer Science. This helps gain insight on how words are formed in the sentences of a lecture material. For example, sentences of "Minumum number of Sentence" < 5 are removed on the preprocessing stage.

### 3.2. Experimental Tools

The following NLP techniques are used for automated generation, manipulation and analysis of data from Figure 1. Data here is the lesson materials from teachers henceforth referred to as documents. In this paper, we represent each sentence in a lesson material as a document. For the auto-generated MCQ's application interface, we used Django. It is a high-level Python Web framework that encourages rapid development and clean, pragmatic design. It also takes care of much of the hassle of Web development and focuses on writing the API without needing to reinvent the wheel

#### 3.2.1. Term Frequency-Inverse Document Frequency

Term Frequency-Inverse Document Frequency (TF-IDF) is used for stop-words filtering in various subject fields including text summarization and classification. The weight is composed of two terms: Term frequency (TF) and Inverse Document Frequency (IDF).

- TF: Term Frequency measures how frequently a term occurs in a document. It is possible that a term would appear much more times in long documents than shorter ones because every document is different in length. Thus, it is stated below as a way of normalization:

$$TF_t = \frac{\text{Number of times term t appears in a document}}{\text{Total number of terms in the document}}$$

- IDF: Inverse Document Frequency measures how important a term is. While computing TF, all terms are considered equally important. However, it is known that certain terms like 'is', 'of', and 'that', may appear a lot of times but have little importance. Thus, it is computed as follows:

$$IDF_t = \text{Log}_e \left(\frac{\text{Total number of documents}}{\text{Number of documents with term t in it}}\right)$$

- The TF-IDFt;d weighting scheme assigns to term t a weight in document d given by

$$TF\text{-}IDF_{t,d} = TF_{t,d} \times IDF_t$$

#### 3.2.2. N-gram

This is a contiguous sequence of n items from a given sequence of text or speech. The items can be phonemes, syllables, letters, words or base pairs according to the application. The n-grams typically are collected from a text or speech corpus. An n-gram of size 1 is referred to as a 'unigram'; size 2 is a 'bigram'; size 3 is a 'trigram'. Larger sizes are sometimes referred to, by the value of n, for example, '4-gram', '5-gram', and so on.

## 4. RESULT DISCUSSION

This discusses the performance of the NLP techniques for MCQs generation applied to the dataset of different lecture materials contents (see Table 1) and compared with the gold standard created by T. Table 3 shows the result of application of the above techniques on documents stated in Table 1 after going through preprocessing in Figure 1. Furthermore, MCQs generation performance scores on the dataset were computed using precision and recall. Precision was to validate the quality of the keywords by MCQs generation extracted from the lecture materials and recall was used to measure the quantity.

Precision answers the question – The keywords extracted from the lecture materials are they important words according to the gold standard marked by the teachers? While recall answers the question - are all the important words in the lecture materials extracted?

Table 3. Statistics of the Keywords automatically extracted from documents.

| Lesson Materials | Number of Keywords Auto Extracted |
|---|---|
| Computer Science | 368 |
| Geology | 318 |
| Philosophy | 254 |
| Religion | 488 |
| History | 471 |

For the performance evaluation, the following metrics precision, recall and f-measure are used in this research.

$$\text{Precision} = \frac{numberCorrect}{numberExtracted} \qquad (1)$$

$$\text{Recall} = \frac{numberCorrect}{totalExtracted} \qquad (2)$$

where numberCorrect is equal to true positive (TP), that is, important keywords extracted. numberExtracted in eqn 1 is summation of TP and false positive (FP). False Positive is extracted keywords that are not important words according to the gold standard. totalExtracted in eqn 2 is the summation of TP and false negative (FN). FN are keywords not extracted but are important words. To find the harmonic mean (average) of the precision and recall for computation of balance between precision (p) and recall (r) in the system:

$$\text{F} - \text{measure} = 2 * \left(\frac{Recall*Precision}{Recall+Precision}\right) \qquad (3)$$

For each lecture material, for instance Computer Science (CS), there are two sets of data:
- Gold standard keywords that are important words in A prepared by the teacher. Henceforth known as truthset.
- The keywords extracted automatically from A. Henceforth known as extractedset.

From the above explanation, assuming truthset and extractedset are sets from CS. TP is a set of intersection between truthset and extractedset (TP = truthset & extractedset), FP is a set of keywords that are in extractedset but not in truthset (fp = extractedset – truthset), while FN is a set of keywords that are in truthset but not in extractedset (fn = truthset – extractedset). Using Table 1, there are 206 truthset and 368 extractedset for CS lesson material.

Table 4. Precision, Recall and F-measure.

| Lesson Materials | TP | FP | FN | Precision | Recall | F-measure |
|---|---|---|---|---|---|---|
| Computer Science | 162 | 206 | 31 | 0.44 | 0.84 | 0.58 |
| Geology | 79 | 293 | 3 | 0.25 | 0.96 | 0.39 |
| Philosophy | 94 | 160 | 2 | 0.37 | 0.98 | 0.54 |
| Religion | 169 | 319 | 1 | 0.35 | 0.94 | 0.51 |
| History | 185 | 286 | 18 | 0.39 | 0.91 | 0.55 |

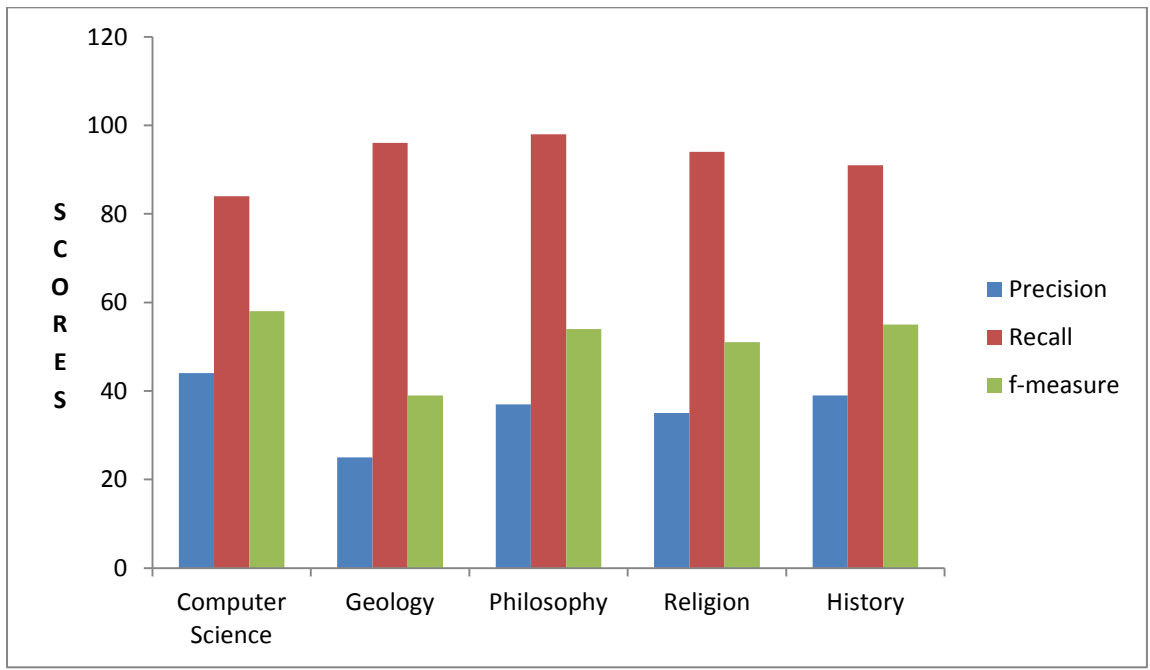

Figure 2. Bar charts of Precision, Recall and F-measure of scores on Table 4.

From Table 4 and Figure 2, it can be observed from the Recall scores that the majority of the keywords automatically extracted are important words. Although Precision scores show that there are other keywords automatically extracted that are not important according to the gold standard set. At the moment we want to ensure that the majority (if not all) the keywords extracted from lesson notes are in the gold standard sets. The keywords extracted by the NLP processes are used to generate Multi-Choice Questions (MCQs).

Since teachers are not technical persons, we develop a user-friendly interface where the MCQs generated are presented for teachers to easily access them. MCQs are generated from the extracted keywords by locating each keyword position given a document.

Table 5. Sample TF-IDF weightd sentences.

| Sentence | TF-IDFWeights |
|---|---|
| The Difference Engine consisted entirely of mechanical components. | {'entirely': 0.305, 'consisted': 0.305, 'components': 0.305, 'mechanical': 0.209, 'difference': 0.209} |

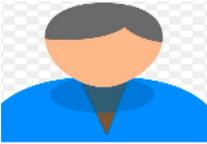

Figure 3. Figure showing API for lecture material upload

.

**question:**

In 1936 , [ Howard ] Aiken had proposed his idea [ to build a giant calculating machine ] to the [ Harvard University ] Physics Department , ... He was told by the chairman , Frederick Saunders , that a lab ________________ , Carmelo Lanza , had told him about a similar contraption already stored up in the Science Center attic .

**options:**

giant

calculating

chairman

technician

**answer:**

technician

Accept

---

**question:**

In 1936 , [ Howard ] Aiken had proposed his idea [ to build a giant calculating machine ] to the [ Harvard University ] Physics Department , ... He was told by the chairman , Frederick Saunders , that a lab technician , Carmelo Lanza , had told him about a similar contraption already stored up in the Science ________________ attic .

**options:**

howard

frederick

Center

university

**answer:**

Center

Accept

Figure 4. Figure showing API for generated MCQs.

Each keyword of a document has a TF-IDF weight resulting in a composite-weighted sentence, this is illustrated in Table 5. How the weight in Table 5 work is when it is highest it means that t occurs many times within a small number of documents (thus lending high discriminating power to those documents); lower when the t occurs fewer times in a document, or occurs in many documents (thus offering a less pronounced relevance signal); lowest when the t occurs in virtually all documents. Given this definition, in a document we extract keywords with top weights as shown in Table 5. These extracted keywords are mapped to their positions within the document so that they are replaced with '–' symbols instead. Then, we blindly and randomly select three other keywords from the pool of extracted keywords plus the main keyword that was replaced with '–' to serve as options from which users are to select the right one. For a document, the number of multi-choice questions (MCQs) to be generated is dependent on the number of keywords extracted (see Table 5). These MCQs are regarded as suggested questions for the teachers to accept or reject. Figure 3 shows the API for lecture material upload while figure 4 show the API for the generated MCQs. The accept button is clicked by the teacher to select an MCQ if satisfied with it.

## 5. CONCLUSION

Multiple-choice questions (MCQs) generation for CBTE using Natural Language Processing (NLP) techniques is a system that helps teachers set multi-choice questions from their documents in a text file and thereafter provide answers to the questions generated. Consequently, it can be used in every area of education sector to test the knowledge or level of the students. However, this paper is not concerned with the issue of whether multiple-choice tests are better assessment methodology than other types of tests. What it focuses on is on application of NLP techniques to generate MCQs about facts explicitly based on keywords as stated in a text format of a given lesson material. From the results, it has shown that the system was capable of extracting keywords from lesson materials in setting examinable questions. This outcome is presented in a user-friendly interface for teachers easy accessibility using Django, a web framework for Python.

## ACKNOWLEDGEMENTS

We thank all the teachers and students that contributed to the success of this work. Your feedbacks are very insightful and helpful towards the development of this work.